\documentclass[letterpaper, 10 pt, conference]{ieeeconf}  

\usepackage{tikz}
\usepackage{textcomp}
\usepackage{hyperref}
\usepackage{lipsum}

\IEEEoverridecommandlockouts                              

\newcommand\copyrighttext{%
  \footnotesize \textcopyright 2024 IEEE. Personal use of this material is permitted.
  Permission from IEEE must be obtained for all other uses. Cite this work: C. Luna et al., "Breadboarding the European Moon Rover System: discussion and results of the analogue field test campaign," 2024 International Conference on Space Robotics (iSpaRo), Luxembourg, Luxembourg, 2024, pp. 145-150, doi: 10.1109/iSpaRo60631.2024.10687700.
  DOI: \href{https://ieeexplore.ieee.org/document/10687700}{10.1109/iSpaRo60631.2024.10687700}}
\newcommand\copyrightnotice{%
\begin{tikzpicture}[remember picture,overlay]
\node[anchor=south,yshift=10pt] at (current page.south) {\fbox{\parbox{\dimexpr\textwidth-\fboxsep-\fboxrule\relax}{\copyrighttext}}};
\end{tikzpicture}%
}

\usepackage{graphicx}
\usepackage{caption}
\usepackage{subcaption} 

\usepackage[
    backend=biber,
    style=ieee,
    sorting=none,
    natbib=true,
    url=false,
    doi=true,
    isbn=false,
    eprint=false,
    maxcitenames=2,
    mincitenames=1,
    maxbibnames=1,
    minbibnames=1,
    autocite=superscript
]{biblatex}
\title{\LARGE \bf
Breadboarding the European Moon Rover System: discussion and results of the analogue field test campaign
}

\author{Cristina Luna$^{1}$, Augusto G\'omez Egu\'iluz$^{1}$, Jorge Barrientos-D\'iez$^{1}$, Almudena Moreno$^{1}$, Alba Guerra$^{1}$, \\Manuel Esquer$^{1}$, Marina L. Seoane$^{1}$, Steven Kay$^{2}$, Angus Cameron$^{2}$, Carmen Cama\~nes$^{3}$, Philipp Haas$^{4}$,\\ Vassilios Papantoniou$^{5}$, Armin Wedler$^{6}$, Bernhard Rebele$^{6}$, Jennifer Reynolds$^{7}$, Markus Landgraf$^{7}$.
\thanks{*The EMRS rover has been developed during European Moon Rover System Pre-Phase A project fully funded by ESA under grant agreement No. 4000137474/22/NL/GLC.}
\thanks{$^{1}$GMV Aerospace and Defence SAU, Calle de Isaac Newton 11, Tres Cantos, Madrid, Spain.
        {\tt\small Corresponding author: cluna@gmv.com}}%
\thanks{$^{2}$GMV NSL Ltd, Airspeed 2, Eighth Street, Harwell Campus, Oxfordshire, UK, OX11 0RL}%
\thanks{$^{3}$AVS Added Value Solutions, Elgoibar, Spain}%
\thanks{$^{4}$OHB System AG, Manfred-Fuchs-Strasse 1, 82234 Wessling, Germany}%
\thanks{$^{5}$Hellenic Technology of Robotics, Kfisias Ave 188, Athens 14562, Greece}%
\thanks{$^{6}$German Aerospace Center (DLR), Institute of Robotics and Mechatronics, Münchener Str. 20, 82234 Weßling, Germany}%
\thanks{$^{7}$ESTEC, ESA, Keplerlaan 1, 2201 AZ Noordwijk, The Netherlands}%
}

\begin{document}

\maketitle
\copyrightnotice

\begin{abstract}

This document compiles results obtained from the test campaign of the European Moon Rover System (EMRS) project. The test campaign, conducted at the Planetary Exploration Lab of DLR in Wessling, aimed to understand the scope of the EMRS breadboard design, its strengths, and the benefits of the modular design. The discussion of test results is based on rover traversal analyses, robustness assessments, wheel deflection analyses, and the overall transportation cost of the rover. This not only enables the comparison of locomotion modes on lunar regolith but also facilitates critical decision-making in the design of future lunar missions.

\end{abstract}

\section{INTRODUCTION}

Humanity has had its gaze set on the stars since an early age. Currently, this vision becomes tangible thanks to advances in space exploration, with cutting-edge missions aimed at human presence on the Moon such as SLIM \cite{EGUCHI2019}, Chandrayaan-3 \cite{pragyan}, Luna 25 \cite{Mitrofanov2021}, Artemis 1 \cite{Hammond2020}, Chang'e 7 \cite{Chi2023}, or Hakuto 1 \cite{Walker2019}, along with their future successors during the current and the following decade.
On Earth, adaptation to different climates and environments has been achieved, enabling the execution of numerous tasks. Now it is time to do so in space. 

This is where the European Moon Rover System (EMRS) comes into play, a project funded by the ESA and developed by an international consortium led by GMV. The goal of EMRS is to create a breadboard of a rover that can adapt to different environments with minimal modifications, enabling it to serve multiple purposes \cite{luna2023european-astra}. EMRS design focus is on modularity \cite{luna2023designing_els, luna2023modularity-iac}. The contribution of this paper lies in discussing the results from the lunar-regolith-analogue test campaign, hereafter referred to as the analogue test campaign, conducted at DLR facilities, evaluating the performance of the multipurpose lunar rover's breadboard. These tests encompass not only the rover's locomotion but also involve the assessment of more complex tasks, such as excavation.

\begin{figure}[t!]
    \centering
    \vspace{6pt}
    \includegraphics[width=\columnwidth]{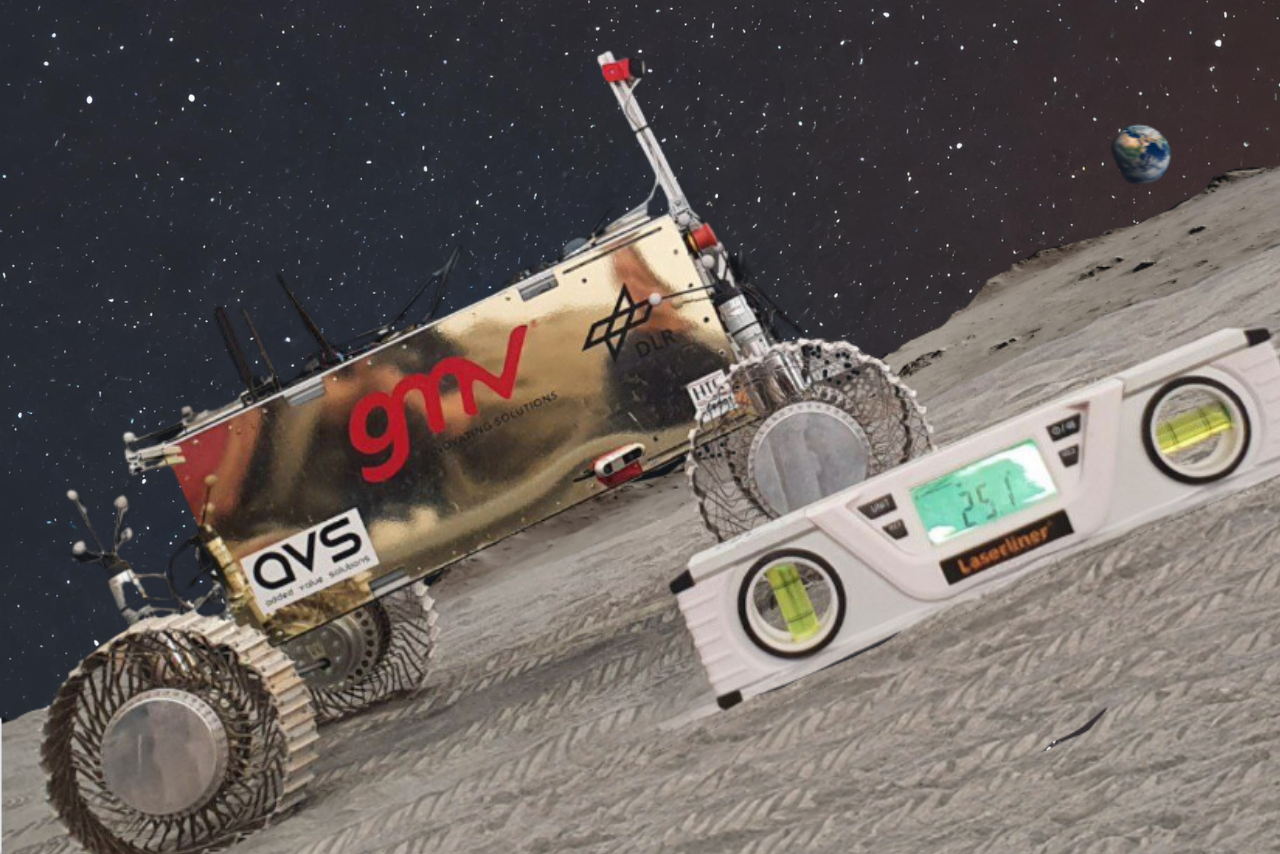} 
    \caption{Conceptual image of EMRS on a crater slope of 25º}
    \label{fig:emrs}
\end{figure}

\section{STATE-OF-THE-ART}

Throughout the development of space exploration, numerous rover models have been sent to the Moon and Mars, alongside others that remain Earth-bound. While some of them have been extensively discussed in a previous work \cite{luna2023modularity-iac}, this section focuses on the most relevant rovers and state-of-the-art developments in lunar rovers .

The VIPER rover, developed by the  National Aeronautics and Space Administration (NASA), is set for a late 2024 lunar mission to explore the Moon's southern pole and investigate frozen water deposits. Weighing 430 kg and measuring 1.53m x 1.53m x 2.45m \cite{viper2020}, VIPER features a one-meter-long regolith and ice drill, Neutron Spectrometer System, Near-Infrared Volatiles Spectrometer, and Mass Spectrometer. Its modular locomotion system, with four modules, enables effective traversal of low-compaction sand at an average speed of 0.2 m/s \cite{viper2020}, showcasing VIPER's versatility in lunar exploration.

Yutu-2, 
developed by the China National Space Administration (CNSA), is
a 140 kg rover with a 20 kg payload, which features instruments akin to Mars rovers. Equipped with panoramic cameras, an infrared spectrometer, and an 
alpha particle x-ray spectrometer
on a robotic arm, it avoids collisions autonomously using sensors. The wheel slip ratios of Yutu-2 \cite{Ding2022}, observed during complex movements, reveal stable performance, with nuanced insights like -0.1 during retractions and 0.1 during uphill maneuvers. These details enhance our understanding of Yutu-2's lunar mobility.

The 
Pragyan
rover, developed by the Indian Space Research Organisation (ISRO), landed on the Moon last year and was designed to operate across uneven terrain \cite{pragyan} for a lunar day.

However, none of these developments focused on modularity and versatility to adapt to different missions. EMRS emphasises the design of a rover characterised by modularity and flexibility, enabling it to support a range of mission configurations.

\section{EUROPEAN MOON ROVER SYSTEM}

The European Moon Rover System (EMRS) Pre-Phase A activity is situated within the European Exploration Envelope Programme (E3P), aiming to provide a versatile surface mobility solution to advance lunar exploration. The study focuses on designing the EMRS with modularity and flexibility \cite{luna2023designing_els} to accommodate various mission configurations while balancing mission versatility and optimality of the system \cite{luna2023european-astra}. The proposed missions include:
\begin{itemize}
\item Polar Explorer (PE): a mission dedicated to exploring water ice deposits on the Moon's polar region. It features a mobile prospecting element equipped with science packages such as the Geophysics Station, the Lunar Surface Environment Package, and the Exobiology Exposure Package.

\item Astrophysical Lunar Observatory (ALO):
This mission aims to elucidate aspects of the deep universe using a low-frequency radio interferometric array. Positioned strategically on the far side of the Moon, it avoids terrestrial interference, enabling observations in the sub-MHz frequency range.

\item In-Situ Resources Utilisation (ISRU): 
a mission of a pilot plant strategically designed to harness lunar resources for sustainable human lunar exploration. Using a stationary pilot plant and a mobile rover to target the Schrödinger Crater for its resource potential, for excavation and resource extraction.

\item Lunar Geological Exploration Mission (LGEM): This mission is part of the Terrae Novae programme and prepares for future lunar science exploration. It features a LIDAR-based instrument onboard the rover to create high-resolution terrain models and provide detailed lunar surface images with a spatial resolution of at least 2 centimetres per pixel.
\end{itemize}

The primary objective of these missions is to contribute to lunar exploration, scientific understanding, and resource utilisation, supporting the broader goals of the European space exploration programme.

\section{EMRS BREADBOARD}

The preliminary mechanical design of the Flight Model (FM) of EMRS was obtained based on mission requirements and design choices, and its details are further explained in \textcite{luna2023european-astra}. From this FM preliminary design, a Breadboard design was derived in order to materialise and test up to TRL 4 the proposed concept.

The breadboard is scaled by a ratio of $1:2$ from the FM, with its main physical characteristics described on Table ~\ref{tab:BB_dims}.

\begin{table}[ht]
\caption{EMRS Breadboard characteristics}
\label{tab:BB_dims}
\begin{center}
\begin{tabular}{|c||c|}
\hline
Main Body Dimensions (L,W,H) & 890 mm, 230 mm, 370 mm\\
\hline
Ground clearance & 250 mm\\
\hline
Distance between wheels (L,W) & 980 mm, 830 mm\\
\hline
Weight & 84 Kg\\
\hline
\end{tabular}
\end{center}
\end{table}

Its structure both of the main avionics box and suspension arms is made of aluminium extrusions, as well as some components of the wheels. For other mechanical assemblies, other materials (mainly stainless steel) are used, such as the elastic elements or their drive trains. Actuation of the wheels is done via $13W$ brushed DC motors with geared speed reduction, all embedded in the wheel hubs. Steering actuation is done with $16W$ brushed DC motors and custom harmonic drive actuator blocks. All these motors are controlled with dedicated speed (for wheels) and position (for steering) closed loop drivers, which are capable of measuring current consumption, thus allowing all the power measurements of the presented tests. 

The Locomotion Software System manages the active joints of the rover (wheels and steering), and performs the accurate kinematics from 2D linear and angular speeds to motor position/speeds according to the desired locomotion mode.

The successful development of the EMRS breadboard validates the modular mobility concept for a multipurpose rover designed for various lunar missions within the European Exploration Envelope Programme (E3P), including PE, ISRU, ALO, and LGEM \cite{luna2023designing_els, luna2023modularity-iac, luna2023european-astra}.

The rover prototype focuses primarily on demonstrating locomotive performance in a representative terrain environment. The fully operational rover, developed using Commercial Off-The-Shelf (COTS) and space components described in \cite{luna2023european-astra}, meets all design requirements \cite{luna2023designing_els, luna2023european-astra, luna2023modularity-iac}. The breadboard has undergone testing in a representative analogue lunar environment, some of the most representative data gathered during the test campaign are described in the following section.

\section{EXPERIMENTAL RESULTS}
\label{sec:exp}

A test campaign to evaluate the functionality and performance of the EMRS breadboard was conducted at the Planetary Exploration Laboratory of DLR, on a slope-configurable sandbox with lunar regolith simulant. The main aspects to be tested were locomotion modes and capabilities to progress through terrain (obstacles and slopes), which are briefly presented in \cite{luna2023european-astra}.
Data gathered from this test campaign relevant to the results shown in the following section include motion capture system information, used as position and orientation ground truth; close-up video recording of two wheels; wheel odometry, and voltage and current consumption on the steering and drive units.

\subsection{Traversing results}
\begin{figure}[h!]
    \centering
    \includegraphics[width=0.5\textwidth]{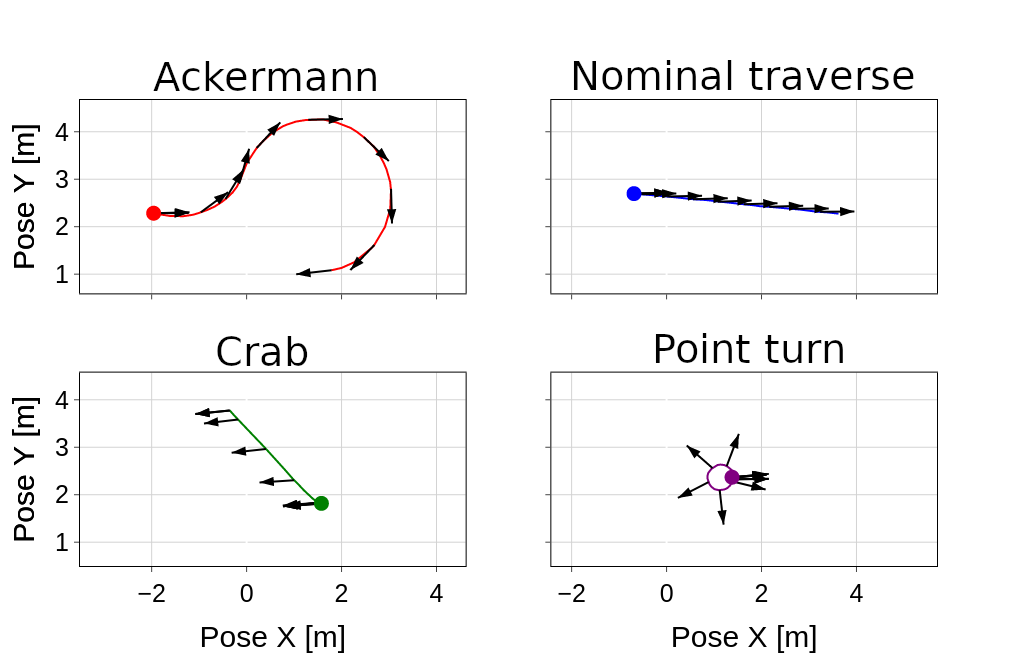} 
    \caption{Trajectories and orientations of the EMRS Breadboard system performing motions in different locomotion modes, as measured by the motion capture system.}
    \label{fig:traverse_plot}
\end{figure}
The locomotion modes evaluated during the testing campaign were Ackermann, Skid, Crab, and Point Turn. These tests were mainly conducted to prove the rover locomotion capabilities, as well as to test the on-board locomotion software. References of linear speed in $X$ and $Y$ axis, and angular speed along the $Z$ axis with the desired locomotion mode were telecommanded to the rover, which transformed them accordingly to wheel speeds and steering positions. Its movement was recorded by the motion capture system, which body marker was positioned at the front part of the rover and slightly rotated respect to the vehicle body axis.

In Fig. \ref{fig:traverse_plot}, the XY plane displacement (horizontal plane) is depicted for each mode, with the rover's orientation represented by arrows at sampled points. In the case of Ackermann steering, the rover executes left and right turns, forming curves until reaching an orientation opposite to the initial one. The orientation vector should be tangent to the curve. For nominal traverse or skid steering, the trajectory appears as a straight line and the orientation arrows align closely with this path. The angle difference between the orientation arrow and the trajectory line results from the rotation of the motion capture marker relative to the vehicle's  axis.

The crab motion test involves a diagonal trajectory with equal velocities along the X and Y axes, resulting in the steering vector and trajectory direction consistently maintaining an angle close to 45 degrees. Finally, the Point Turn test exhibits displacement on the horizontal plane due to the distance between motion capture marker and the centre of the vehicle. Additionally, the directional vector consistently points outward from the circumference.

\subsection{Steering modes analysis}

The proposed EMRS design has individual steering for all four of its wheels. These steering units add cost, system complexity, possible points of failure and other contingencies such as thermal management of its actuators that could imply extra power needs. Considering this, the necessity of individual steering should be assessed and justified when designing a rover. 

\begin{figure}[h!]
    \centering
    \includegraphics[width=0.45\textwidth]{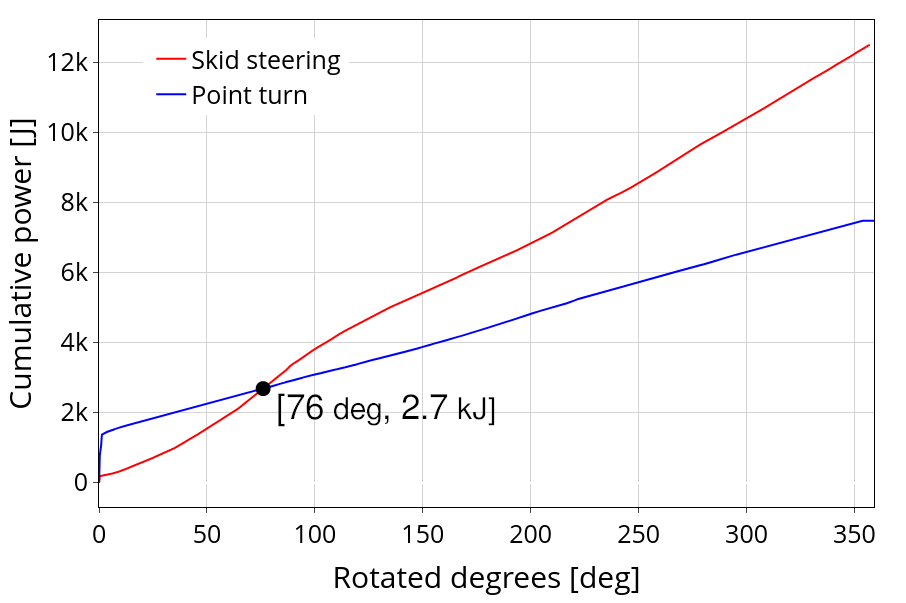} 
    \caption{Energy consumption per degree of rover yaw captured from ground truth. Comparison between skid-steering and point turn rotations.}
    \label{fig:rotation_energy}
\end{figure}

A comparison of two pure rotation modes, one without steering modules actuation (\textit{skid steering}) and one with all four steering modules in use (\textit{point turn}). The first comparison is shown in Fig. \ref{fig:rotation_energy} in terms of power consumption per yaw angle. \textit{Point turn} requires an initial actuation of the steering actuators to point all the wheels axis of rotation to the centre, which implies a basal energy consumption that \textit{skid steering} does not have. However, as yaw angle increases, the better efficiency of \textit{point turn} steering results in overall lower energy consumption.

\begin{figure}[h!]
    \centering
    \includegraphics[width=0.45\textwidth]{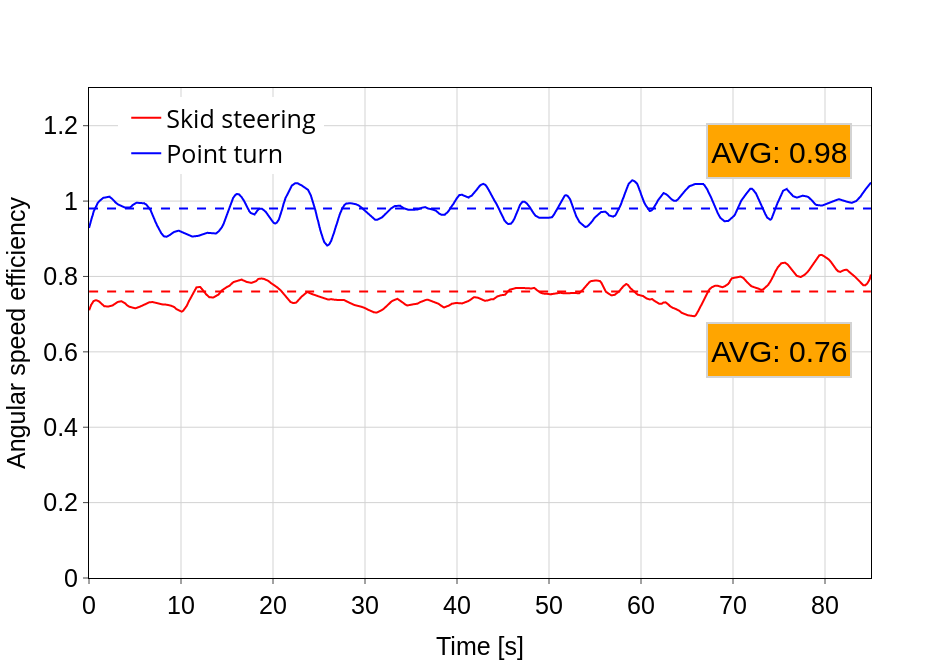} 
    \caption{Angular velocity ratio between the obtained with ground truth and through wheel odometry over the time.}
    \label{fig:angular_speeed_efficiency}
\end{figure}

The loss of wheel motion due to skidding not only implies worst energy usage but has even further implications in the rover system. Fig. \ref{fig:angular_speeed_efficiency} shows the angular speed efficiency defined as the measurement obtained by the ground-truth system divided by that calculated using wheel odometry. The angular speed estimated through wheel odometry in \textit{point turn} against the one obtained by motion capture system bears little error, so the efficiency is approximately $100\%$. Conversely, the error between the ground-truth speed and the one obtained through \textit{skid steering} odometry yields a loss of wheel motion of approximately $25\%$, which is considerable, but also expected as the lunar regolith simulant has a very low compaction. Furthermore, this combination of skidding, regolith characteristics, and wheel geometry caused the wheels to carve themselves into the soil, risking the rover getting stuck and loosening the regolith, which in turn led to more frequent landslides when tested on the steeper slopes ($15^\circ$ and $20^\circ$).

It is worth noting that these results are particular to the presented rover breadboard. However, they should be taken into consideration when designing a lunar rover according to the overall system requirements, i.e. cost, concept of operations and estimated lifespan, among others.

\subsection{Wheel deflection analysis}

Understanding wheel deflection is a critical factor for future developments of EMRS, such as calculating the traction of the rover during excavation tasks or suspension fine tunning to handle larger payload capacities.
In this analysis, the deflection of the rover wheel while moving at $0.6$ metres per second, clearing a $0.16$ m. obstacle without extra payload was studied. This case involved one wheel overcoming an obstacle while the others stay on the ground. Overcoming the obstacle under these conditions entails that, in addition to the wheels moving on the ground, there is also a certain moment when one wheel is in the air sharing the rover's weight among the other three wheels. 

\begin{figure}[h!]
    \centering
    \begin{subfigure}{.3\textwidth}
      \centering
      \includegraphics[width=1\linewidth]{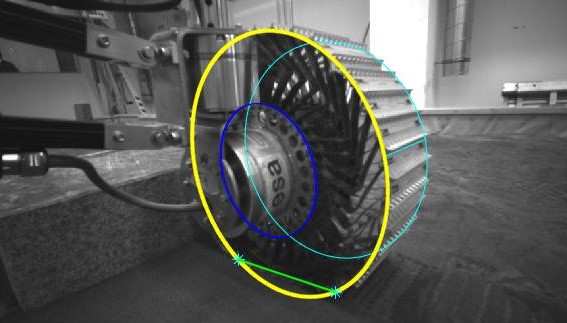}
    \end{subfigure}
    \begin{subfigure}{.165\textwidth}
      \centering
      \includegraphics[width=1\linewidth]{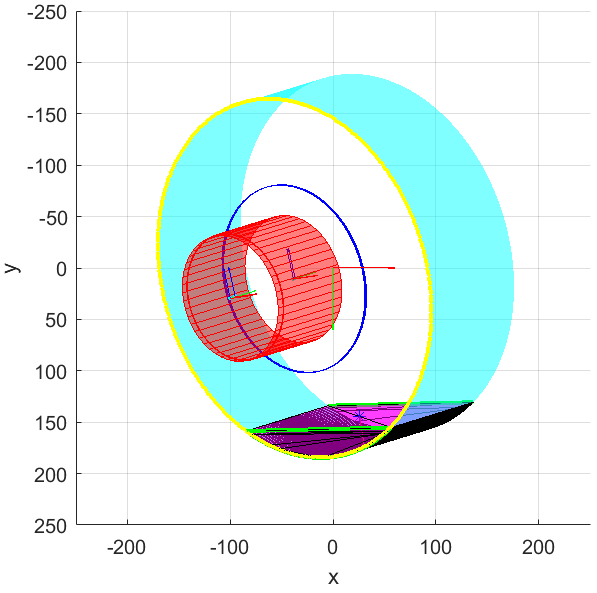}
    \end{subfigure}
    \begin{subfigure}{.3\textwidth}
      \centering
      \includegraphics[width=1\linewidth]{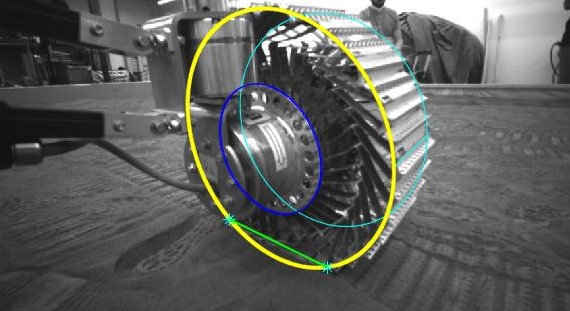}
    \end{subfigure}
    \begin{subfigure}{.165\textwidth}
      \centering
      \includegraphics[width=1\linewidth]{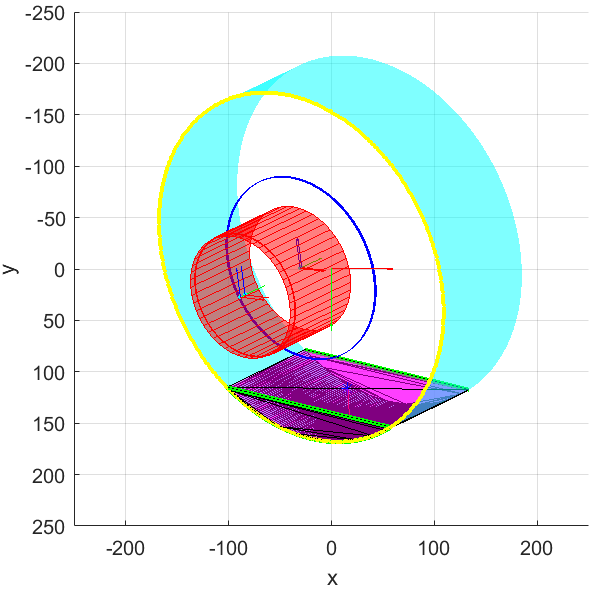}
    \end{subfigure}
    \caption{Wheel deflection estimation for each of the two wheels A and B on the top and bottom, respectively. On the left, image annotation of wheel pose reprojection and deflection. On the right, 3D model of the wheel and deflected volume estimation (in magenta).}
    \label{fig:wheel_image_reproj}
\end{figure}

Images of the wheels were obtained during the experiment from their corresponding calibrated cameras. For simplicity, we will denote hereafter as \textit{wheel-A} and \textit{wheel-B} to refer to the right-front and left-back wheels. A semi-supervised labelling procedure was performed to annotate the correct positioning of the wheels and their deflection at each frame. A simplified 3D model of the wheel was used to estimate its pose with respect to the camera by fitting its reprojection on the image plane, i.e. using the camera intrinsic calibration parameters. 
Fig. \ref{fig:wheel_image_reproj} shows an example of the image annotation and corresponding 3D model for both wheels (i.e. \textit{wheel-A} and \textit{wheel-B}) at the top and bottom, respectively. 
The 3D model consists on three circles that represent the inboard and outboard perimeter (in yellow and cyan) and the wheel hub plate (in blue). The wheel deflection at each frame was approximated as a 2D line (in green) that undercuts the internal wheel perimeter in the image. The line representing the wheel deflection is initially estimated using a Hough Transform approach on Canny filtered grayscale images and, when needed, manually refined. The intersections of the line with the inboard perimeter reprojection circle fit were computed, and their projection on the 3D space was used to estimate the deflected part of the outboard perimeter by assuming equal deflection on both sides of the wheel.
A convex hull of the points laying on the inboard and outboard perimeters between the four intersection points was computed, and the resulting polygon's volume (shown in magenta) was assumed to be the wheel deflection at the time the frame was captured. As the volume of the wheel with no deflection is known from its geometry, the relative wheel deflection at each frame was estimated. 


\begin{figure}[!h]
    \centering
    \includegraphics[width=0.4\textwidth]{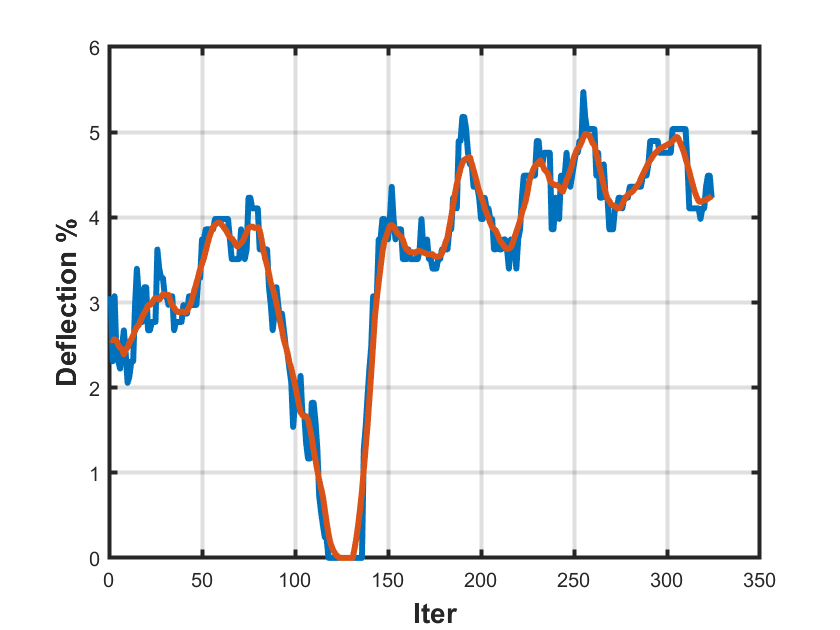} \includegraphics[width=0.4\textwidth]{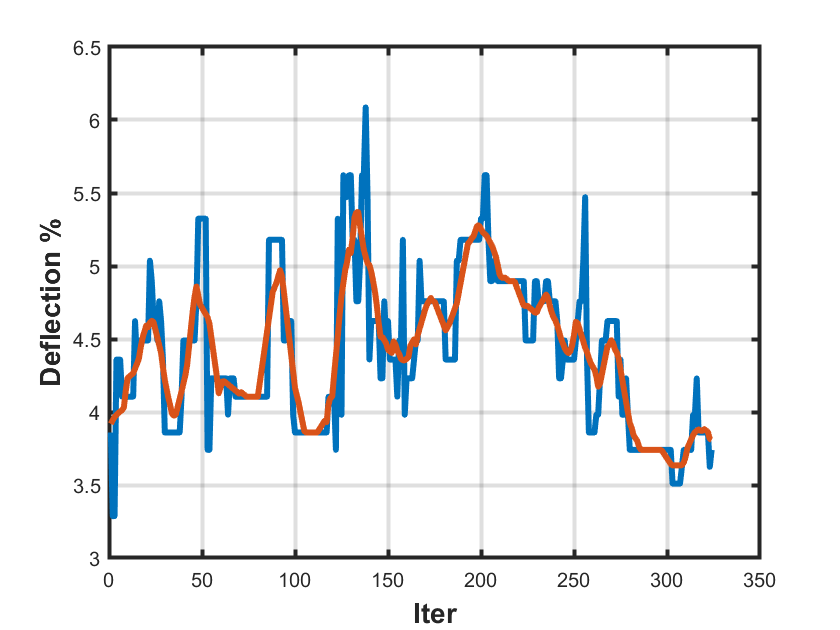}
    \caption{Deflection of wheels A and B on the left and right, respectively.}
    \label{fig:wheel_AB}
\end{figure}

Fig. \ref{fig:wheel_AB} shows the evolution of \textit{wheel-A} and \textit{wheel-B} deflected volume over time during the experiment. The deflected volume was computed by dividing the estimated volume of a wheel at a given time step by its original total volume. The x-axis corresponds to the image frames, and the y-axis corresponds to the percentage of wheel deflected. The orange line represents a smooth function of the wheel deflection estimates, and the blue line represents the raw estimates. The experiment begins with \textit{wheel-A} raised over an obstacle, leading to an unbalanced rover weight distribution. Between frames $1$ and $90$, \textit{wheel-A} is still flat on the obstacle. Then, it jumps out of the obstacle, progressively returning to its shape as the contact with the obstacle is being reduced. Between frames $125$ and $137$, \textit{wheel-A} is in the air and, therefore, the wheel has not deflected. After the impact, the wheel deflection returns to a range similar to before getting off the obstacle. Then, it stabilises around $3.5$ and $5$\% as a consequence of re-balancing the rover’s weight among all the wheels. Regarding \textit{wheel-B}, between frames $1$ and $137$ (i.e. the impact of \textit{wheel-A} after falling down from the obstacle), the wheel deflection is stable between $\sim3.5$ and $\sim5$\%. Interestingly, the greater peak can be observed at the impact of \textit{wheel-A} with the ground, indicating that the rover’s weight is mostly kept on \textit{wheel-B} at that instant. As the suspension compression modifies the suspension’s weight distribution, the impact with the ground of \textit{wheel-A} is reflected in \textit{wheel-B} deflection. 
Then, similarly to \textit{wheel-A}, the \textit{wheel-B} deflection returns to a range similar to before getting off the obstacle.

It is worth noting that the EMRS rover is a breadboard, and the wheels are stiffness configurable. The deflection of the wheels might slightly differ from one another under the same conditions, and therefore, there is not enough evidence to state a nominal wheel deflection value for all wheels. However, in conclusion, experimental data have shown the nominal wheel deflection for the EMRS rover is within the range of $\sim3.5$ and $\sim5$\%. It was observed that the suspension and weight distribution also play a role in the deflection of each wheel, which stayed at all moments (i.e. even after an impact with the ground) below $6.5$\%.

\subsection{Cost of transport}
A series of tests were conducted during the campaign consisting on the ascent of a ramp with varying inclination in a linear trajectory. These tests were undertaken with the dual purpose of demonstrating the capability of the vehicle to moving at different speeds, ascend ramps or excavate and to measure its energy consumption. The inclinations examined ranged between $0^\circ$ and $25^\circ$.

The assessment of this energy consumption performance is done via the Cost of Transport formula (\ref{eq:CoT}) derived from the work of \textcite{gabrielli1950price}. Cost of Transport is a non-dimensional magnitude used to measure the energy cost associated with moving a certain mass over a specific distance. It is often used in the context of transportation systems, particularly in fields such as biomechanics, robotics, and vehicle engineering.
Thus, the Cost of Transport is defined as:
\begin{equation}
    \epsilon = \frac{P}{mgv}
    \label{eq:CoT}
\end{equation}
where $\epsilon$ is the cost of transport, $P$ is the power consumed in $[W]$, $m$ is the mass of the rover in $[kg]$, $g$ is the gravity ($[m/s^2]$) and $v$ is the linear speed in $[m/s]$ \cite{Trancossi_2015}.

Table \ref{tab:tab-cot} depicts the Cost of Transport relative to the different test. To achieve this, the mean cost was computed for each test conducted at varying inclinations, including flat surface and excavator test. The speed used is the one obtained directly from the motor encoders, after analysing the slippage with the delta between this speed and the speed calculated by the motion capture system.

\begin{table}[h]
    \centering
    \caption{Cost of transport.}
    \begin{tabular}{|c|c|c|c|}
        \hline
        \textbf{Mode} & \textbf{Degrees of slope} & \textbf{Velocity}  & \textbf{Cost of transport} \\
        \hline
        Excavator & 0 & 3 cm/s & 0.553\\
        \hline
        Nominal & 0 & 3 cm/s & 0.646\\
        \hline
        Nominal & 0 & 6 cm/s & 1.10\\
        \hline
        Nominal & 0 & 8 cm/s & 1.39\\
        \hline
        Slope up & 10 & 6 cm/s & 0.891\\
        \hline
        Slope up & 15 & 6 cm/s & 0.769\\
        \hline
        Slope up & 20 & 6 cm/s & 0.591\\
        \hline
        Slope up & 25 & 6 cm/s & 0.694\\
        \hline
    \end{tabular}
    \label{tab:tab-cot}
\end{table}

The EMRS rover is designed to be able to carry different payloads and weights as well as to perform many different tasks in uneven terrain. This implies that the efficiency point of the breadboard wheel motors is reached once the rover is loaded or the rover is traversing difficult terrain, which is expected on the surface of the Moon, not flat surfaces at small speeds. The scaling of the size of the rover, as well as its components, has been carried out following an approach explained in \cite{luna2023european-astra}; which means that the speeds used in the tests are slightly lower than those expected in the flight design (an average speed of 12.67 cm/s).

\section{DISCUSSION}

The EMRS Breadboard's development validates the modular mobility concept for a versatile lunar rover. Employing Commercial Off-The-Shelf (COTS) and space components, the prototype adheres to design requirements and undergoes testing in a representative analogue lunar environment, aligning with its intended adaptability for various lunar missions.

In Section \ref{sec:exp}, locomotion capabilities in Ackermann, Skid, Crab, and Point Turn modes are evaluated. The trajectories and orientations, illustrated in Fig. \ref{fig:traverse_plot}, showcase distinct behaviours. Notably, the rover has surpassed the requirements in executing these diverse locomotion modes, underscoring its robust performance. This achievement is attributed to the modular design, which enables adaptability and versatility in locomotion \cite{luna2023european-astra}. The choice of locomotion type is now contingent on the specific operational requirements of the mission, highlighting the significance of the rover's modular design in tailoring its mobility to varying scenarios.

The analysis transitions to individual steering modes for the four wheels, introducing additional costs and complexities. Comparing Skid-Steering and Point Turn modes exposes nuanced energy dynamics. Despite the higher basal energy in Point Turn due to initial steering actuation, its superior energy efficiency with increasing yaw angle is evident (Fig. \ref{fig:rotation_energy}). These findings suggest that while Point Turn and Ackermann steering modes are generally more energy-efficient, there are subtleties that need to be considered. Depending on the mission's operational requirements, it may be strategically beneficial to employ Point Turn or even Skid-Steering in specific scenarios, taking into account factors such as terrain mechanics and the interaction between the rover and the lunar surface.

Wheel deflection during obstacle traversal is analysed using images for semi-supervised annotation (Fig. \ref{fig:wheel_AB}). Nominal wheel deflection, approximately 3.5\% to 5\%, is observed, stabilising post-obstacle traversal. However, it is essential to note that due to the inherent design constraints of the prototype, ensuring identical stiffness settings for both wheels was not feasible. Consequently, a direct comparison with analytical models of wheel sinkage \cite{Zhu2023} could not be established. This limitation emphasises the potential impact of wheel stiffness configuration on observed variations, highlighting the need for further investigations to precisely quantify this aspect in future lunar rover studies.

The test analysis wraps up by evaluating energy consumption during ramp ascents  with different inclinations, flat traverses and excavator test, using the Cost of Transport formula (\ref{eq:CoT}) \cite{gabrielli1950price}. In general, the prototype rover demonstrates remarkable efficiency in terms of mission cost of transport, making it well-suited for operations in highly variable environments. The test were conducted using the basic EMRS breadboard without lateral and top modules designed to accommodate different payloads, decreasing the weight of the rover and hence, the weight expected in some tasks as discussed in the cost of transport section. Table \ref{tab:tab-cot} shows the calculated CoT and depicts the suitability of the EMRS rover to carry different tasks in different scenarios and the versatility of its design.

These results provide an insight of the EMRS Breadboard's abilities. The significance of these experimental discoveries is evident in their impact on future developments in lunar rover technology, considering important factors like expenses, operational dynamics, and expected lifespan.

\section{CONCLUSIONS AND FUTURE WORK}

In this paper, a modular rover designed for the EMRS project is presented. The focus is on providing detailed results of the test campaign to validate the EMRS rover for lunar exploration. The rover's unique modular design enables the evaluation of locomotion, software, and various scientific payloads simultaneously. This approach enhances understanding of the rover's capabilities and offers valuable insights for the development of future lunar rovers. The project's success highlights the significance of adaptability in tackling dynamic challenges. The lessons learned from this project are expected to have a significant impact on the design and implementation of future lunar rovers, ensuring effective, efficient, and long-lasting exploration of the lunar surface. The EMRS team is currently engaged in future work, focusing on adapting and extending the EMRS rover to follow a path-to-fly and adapting the concept to smaller missions. Additionally, efforts are underway to enhance mobility capabilities through adding active suspension capabilities and addressing different egress and payload operations.

\section*{ACKNOWLEDGEMENTS}

EMRS highlights successful collaboration among European companies, showcasing a versatile rover developed collectively with the support of ESA. Gratitude to GMV, AVS, HTR, OHB, and DLR for their dedicated efforts, and appreciation to ESA for continuous support to the European space industry and their support to us during the EMRS project.


\AtNextBibliography{\small}
\printbibliography

\addtolength{\textheight}{-12cm}

\end{document}